\documentclass[10pt,letterpaper]{article}

\usepackage{cogsci}
\usepackage{pslatex}
\usepackage{apacite}
\usepackage{graphicx}
\usepackage{booktabs}
\usepackage{xcolor}
\usepackage{amsfonts}
\usepackage{tablefootnote}

\newcommand{\quash}[1]{} 

\title{A Nurse is Blue and Elephant is Rugby: \\ Cross Domain Alignment in Large Language Models Reveal Human-like Patterns}

\author{Asaf Yehudai$^*$ ~~~ Taelin Karidi$^*$ ~~~ Gabriel Stanovsky ~~~ Ariel Goldstein  ~~~ Omri Abend
\vspace{0.2cm} \\
Hebrew University of Jerusalem \\
\small{\texttt{\{asaf.yehudai, taelin.karidi, gabriel.stanovsky, ariel.goldstien, omri.abend\}}@mail.huji.ac.il}}

\begin{document}

\maketitle

\begin{abstract}

Cross-domain alignment refers to the task of mapping a concept from one domain to another. For example, ``If a \textit{doctor} were a \textit{color}, what color would it be?''. This seemingly peculiar task is designed to investigate how people represent concrete and abstract concepts through their mappings between categories and their reasoning processes over those mappings. 
In this paper, we adapt this task from cognitive science to evaluate the conceptualization and reasoning abilities of large language models (LLMs) through a behavioral study. We examine several LLMs by prompting them with a cross-domain mapping task and analyzing their responses at both the population and individual levels. Additionally, we assess the models' ability to reason about their predictions by analyzing and categorizing their explanations for these mappings. 
The results reveal several similarities between humans' and models' mappings and explanations, suggesting that models represent concepts similarly to humans.
This similarity is evident not only in the model representation but also in their behavior. Furthermore, the models mostly provide valid explanations and deploy reasoning paths that are similar to those of humans.


\quash{\textbf{Keywords:} 
Cross-Domain Mapping; Conceptualisation; Large Language Models; Concreteness; Associations}

\renewcommand{\thefootnote}{\fnsymbol{footnote}}
\footnotetext[1]{Equal contribution.}
\renewcommand*{\thefootnote}{\arabic{footnote}}

\end{abstract}

\section{Introduction}
\label{intro}

Large Language Models (LLMs) have significantly improved their ability to generate human-like text and tackle complex tasks that require reasoning. However, the ability to explain or present their behavior in human-understandable terms has remained a challenge \cite{doshi2017towards,du2021towards,zhao2023explainability}.
The ability to interact with them in a very similar way to humans, has encouraged researchers to evaluate their understanding and reasoning abilities by comparing their behavior to humans \cite{futrell2019neural, Binz_2023}, drawing insights from fields such as cognitive psychology, psycholinguistics, and neuroscience \cite{huth2016natural,pereira2018toward,futrell2019neural,goldstein2022shared}.

\begin{figure}
   \centering
   \includegraphics[width=0.8\linewidth]{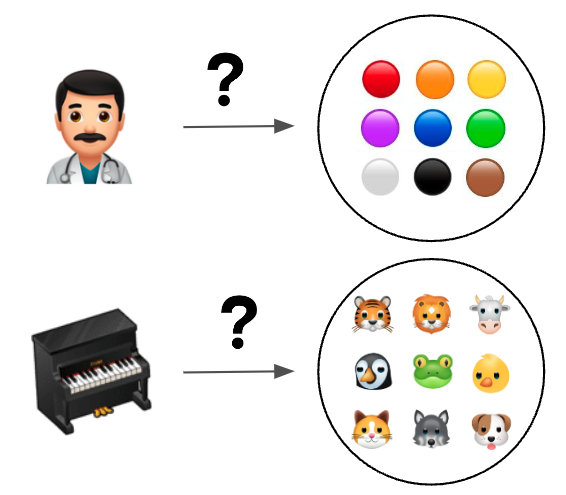}
   \caption{Cross-Domain Alignment: Mapping object from domain A to domain B. Here, doctor from the profession domain to the color domain, and piano from the instrument domain to the animal domain.}
   \label{fig:fig_1}\vspace{-1em}
\end{figure}

In this paper, we perform a behavioral study on LLMs, drawing inspiration from a recently created psychological task that has claimed to uncover aspects of how people represent concrete vs. abstract concepts, and the conceptual organization of metaphoric language \cite[henceforth {\it LL23}]{liu2023cross}.
In this task, participants are asked to map concepts from one semantic domain to another (e.g., \textit{doctor} to \textit{color}, and \textit{piano} to \textit{animal}, see Fig. \ref{fig:fig_1}) and explain their choices. 
Interestingly, it was shown that individuals perform these seemingly arbitrary mappings in predictable ways, relying on certain types of similarity such as \textit{perceptual similarity} or \textit{word associations}. For example, \textit{drum} was consistently mapped to \textit{thunder}, clearly motivated by their \textit{sensory similarity}, as they both make a similar noise. Understanding the basis of these mappings offers insights into the participants' conceptual representation and organization, similar to how psychologists use mental associations or nonsensical tests to assess human conceptual representations \cite{greenwald1998measuring,davis2019does}. 
We further investigate the reasoning behind their mappings by analyzing their explanations of the mapping responses, making them more interpretable.

Our experiments are divided into two main parts: (1) Cross-domain mapping task, and (2) Explanations of the mappings.
We first ask whether LLMs can perform cross-domain mappings and if so, whether they converge with human behavior. To answer these questions, we use human data collected as part of cognitive experiments, to prompt several LLMs with cross-domain mappings.
Surprisingly, results show that models can perform such mappings, reaching a substantial agreement at the population level, which is much higher than a random chance guess.
Moreover, some LLMs surpass the individual level agreement with the population level (i.e. most popular) mappings, showing that their behavior is closer to the ``typical'' human behavior than that of a random participant.

To further interpret the models' mappings and their ability to reason about them, we prompt the models to explain the mappings.
We use the predefined similarity categories (e.g., \textit{perceptual similarity}) that were found to establish the basis of alignment for humans and train a classifier to classify the models' explanations according to them. We see that the models' explanation categories are distributed in a very similar way to humans, suggesting they rely on similar types of similarity in their representations.
Moreover, we perform a qualitative analysis of their explanations, showing they can give concise arguments for the cross-domain mappings. 
Our findings can contribute to the recent discussion in the NLP and cognitive literature as to whether we can ascribe concepts such as conceptualization to LLMs, by demonstrating that at least at the behavioral level, alignment can be found between LLMs and humans in conceptualization tests \cite{bubeck2023sparks,kosinski2023theory,binz2023turning}.


\section{Background}
\label{sec:back}

In recent years, the use of methods and experiments from the cognitive psychology literature in the field of NLP, has become more ubiquitous \cite{dasgupta2020theory,hagendorff2023human,ullman2023large}. LLMs are also used as cognitive models, as they offer almost accurate representations of human behavior, even outperforming traditional cognitive models on some tasks \cite{binz2023turning, suresh2023conceptual,goldstein2023decoding}. 
Moreover, it was shown that improving model-human alignment on psychological tasks results in model improvement on various downstream applications \cite{sucholutsky2023alignment}. 


Recent psychological research proposes a new task, cross-domain alignment, to specifically investigate concepts' semantic similarity between different domains (LL23).
Cross-domain alignment refers to the task of mapping a concept from one domain to another (LL23). Comparing words (concept) within a \textit{semantic domain} (e.g. \textit{dog/cat} or \textit{nurse/doctor}) is relatively easy as they have a multitude of common features, and tend to share similar functions. However, it is less intuitive to align concepts from different semantic domains -- such as \textit{nurse} (profession) to \textit{blue} (color) or \textit{guitar} (instrument) to \textit{rain} (weather).
The research concludes that even concrete concepts are mentally organized along more abstract dimensions. An example of such a dimension is valence; experiments indicate that individuals tend to assign positive or negative valence to concrete concepts in domains like colors, professions, and beverages. Here, we use this task to get a glimpse into LLMs' conceptualization and reasoning abilities.

\section{Cross-Domain Mapping}
\label{sec:cross_dom_map}

In this section, we describe the cross-domain mapping experiments. We perform two types of analysis -- population-level analysis, in which we compare the model's behavior to a ``typical'' human behavior. We then perform an individual-level analysis where we compare the model's behavior and a single participant's behavior.

\subsection{Experimental Setup}
\label{subsec:exp_setup}


\paragraph{Dataset.} For our experiments, we use the cross-domain mapping data collected from humans by LL23 in their experiments (full details are in the Appendix).
The data contains 12 domains\footnote{The domains were initially selected in the original paper, LL23.} (see Table \ref{app:tab:categories_concepts} in the Appendix), from which 32 domain-pairs were selected. For each domain pair, 2-3 random statements of the form:``If a(n) $x$ (source item) were a(n) $y$ (target domain), what $y$ would it be?'' (e.g., ``If a \textit{doctor} (source item) were a \textit{color} (target domain), what color would it be?'') are constructed. Resulting in 75 statements, each answered by 20 participants. 

\begin{figure*}[t!]{}
\centering
\includegraphics[width=\textwidth]{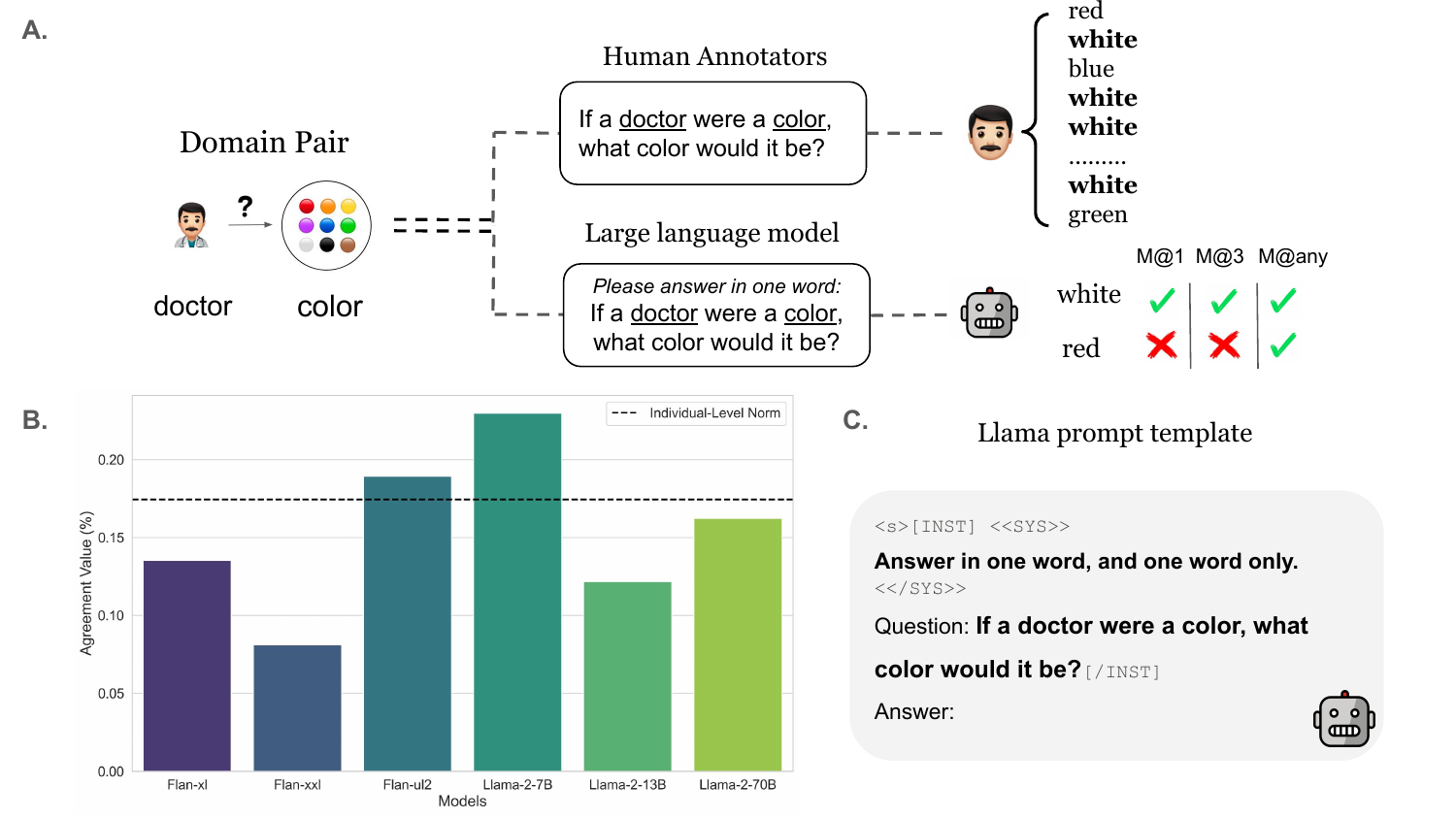}
\caption{\textbf{A.} An illustration of our evaluation pipeline, at the top, is the human annotation process of LL23, and at the bottom is our LLM evaluation process. \textbf{B.} Model-human Agreement: each bar represents the M@1 score. The dashed line represents the individual-level norm for agreement with the most popular answer. \textbf{C.} Llama prompt template. An example of a Llama prompt template, for the domain-pair \textit{(doctor, color)}.}
\label{fig:method}\vspace{-1em}
\end{figure*}

\paragraph{Models.} We select seven robust LLMs, including variants of Flan language models \cite{wei2022finetuned} and Llama-chat language models \cite{touvron2023llama} and Mistral-7B\footnote{Mistral is the only model we use for generating explanations for the received cross-domain mappings. This decision is based on the observation that Mistral does not follow the task's instructions, such as providing responses that are not concise or relevant to the task (this may be attributed to its training and alignment process).} (See Appendix for full details).  
These models, termed ``instruction-following LLMs,'' belong to a category of language models specifically trained to follow instructions -- an important trait in our context. 
We select these models for their accessibility and their high performance. 

\paragraph{Prompt Templates.} Following recent work, to reduce the noise in the models' response, we manually construct 4 templates to prompt the models \cite{mizrahi2023state, rabinovich2023predicting}. For example, ``If a(n) $x$ (source item) were a(n) $y$ (target domain), it would be a(n) '' (see Appendix for a full description of the templates). We choose the indefinite article preceding $x$ and $y$ to construct a grammatical sentence. For the Llama models that are oriented towards more lengthy conversational responses, we adjust the meta-prompt to encourage them to provide short answers that fit the format.

\paragraph{Prompting Methodology.} Prompting the model with 4 templates per statement, results in 4 responses. We use \textit{majority vote} to consolidate the results, and have one model response per statement.\footnote{In case of a tie we randomly choose one response.} We use greedy decoding as a way of approximating the model's most probable response. Notably, different templates and different decoding schemes can lead to different responses that in turn affect the model's behavior. However, these interventions were designed to ensure responses that better capture the model's predominant behavior. 

\paragraph{Metrics.}
To score the model's performance we use Match at K metric, {\it M@K}, i.e., we check if the model's answer is within the first K most popular human answers. Formally, we denote the model's responses by $\{r_1,\ldots,r_{N}\}$ and the humans' responses by $\{h_1^{j},\ldots,h_{N}^{j} \}$ where $N$ is the number of statements, $N = 75$, and $j$ is the popularity of the answer, with 1 representing the most popular answer. Accordingly, M@K is defined as the average over the indicator ${\delta}_{ij}$ representing if for statement $i$, the model's response $r_i$ is the same as the human response $h_i^j$. For our experiments, we will use, M@1, M@3, and M@any, with M@any indicating that the model's response is one of the humans' answers.

We note that 20 human participants are a relatively small group to establish responses at the population level, nevertheless, LL23 found responses to converge in a nontrivial way, indicating these responses are at least somewhat representative. To overcome this we chose to incorporate M@any and asses if the model answer is reasonable even if less probable.

\subsection{Experiments $\&$ Results}

Since there are no clear-cut ``right'' or ``wrong'' responses for cross-domain mappings (a characteristic typical to any projective test), we conduct our analysis at both the population and individual levels. This allows us to assess LLMs behavior by comparing it to human behavior.\footnote{Our code and supplementary materials are publicly available at \url{https://github.com/tai314159/XCategory}}

\paragraph{Population-Level Analysis.}

We begin by evaluating the capability of LLMs to perform cross-domain mappings, comparing their performance to that of 'typical' or average human behavior
For each domain pair, we have one response from the model and $\sim 20$ responses from human participants.  
We then compute M@1, M@3 and M@any.

\paragraph{Results.}
Table \ref{tbl:cda_scores} presents the M@1, M@3, and M@any for the Flan and Llama models. The M@1 score ranges between $8.1\%-24.3\%$, and the M@3 ranges between $14.9-36.5$. Llama-7B scores the highest in both M@1 and M@3 with $24.3\%$ and $36.5\%$ respectively. The M@any shows further increase and has higher than $50\%$ of being a human answer for most models. It is easy to be convinced that these scores are much higher than a random chance baseline of just guessing a concept from the target domain.

Interestingly, we see that larger models do not necessarily score higher. A possible reason for this might be that their responses are less similar to those of humans, but still acceptable or even semantically equivalent (e.g., answering \textit{latte} where humans answer with \textit{coffee}). Verifying this requires the definition of a metric that assesses semantic equivalence rather than word-level matching, a direction we leave for future work.
Another explanation can be the size of the training data.\footnote{Flan-ul2 was pre-trained on 1 trillion tokens, Llama was trained on 2 trillion tokens, Flan-xl and -xxl were trained for at most 350B tokens, based on the T5, and T5-LM-adapt versions \cite{raffel2023exploring}.}
We note that as the evaluation set is relatively small these conclusions warrant further investigation. Nonetheless, our main claim is that LLMs perform cross-domain mappings at an above-chance level,\footnote{The chance level for any response (which is uniquely defined by a domain-pair) can be interpreted as the probability of randomly choosing a concept from the target domain.} resembling the patterns observed in human cognitive processes across all models.

\begin{table}[t]
    \centering
    \resizebox{0.85\columnwidth}{!}{%
        \begin{tabular}{@{}lccc@{}}
            \toprule
            Model Name & M@1 & M@3 & M@any \\ 
            \midrule
            Flan-xl                        & 12.2 & 23.0 & 39.2   \\
            Flan-xxl                       & 8.1  & 14.9 & 32.4   \\
            Flan-ul2                       & 18.9 & 28.4 & 55.4   \\
            Llama-7B                       & \textbf{24.3} & \textbf{36.5} & 54.1   \\
            Llama-13B                      & 21.6 & 27.0 & \textbf{56.8}   \\
            Llama-70B                      & 16.2 & 29.7 & 50.0   \\ 
            \bottomrule
        \end{tabular}%
    }
    \caption{Cross-Domain Alignment scores for several models. Here, M@n indicates that the model answer is within the top n most popular human answers. 
    }
    \label{tbl:cda_scores}\vspace{-1em}
\end{table}

\paragraph{Individual-Level Analysis.}

Our task can be categorized as a \textit{projective test}, similar to the Word Association and Rorschach tests, in which the patient is expected to project unconscious perceptions by the test stimuli. In such tests, the popular answers at the population level are not expected to be produced at the individual level. 
To further understand the behavior of LLMs compared to humans, beyond just the average human behavior, we additionally conduct an analysis at the individual level.

In our task, due to the absence of predefined individual behaviors in the experimental data, we employ a manipulation at the single annotator level to infer individual-level norms for cross-domain mappings. To this end, we iteratively exclude one annotator from the pool of annotators and score their answers compared to the popular answers defined by the rest of the annotators. This procedure allows us to establish individual-level norms, however, as we sample from the distribution defined by the annotations we can see this norm as an upper-bound to the actual individual-level norm. Nonetheless, this individual-level norm facilitates a comparison between individuals and LLMs. 

\begin{figure*}[t!]{}
\centering
\includegraphics[width=\textwidth]{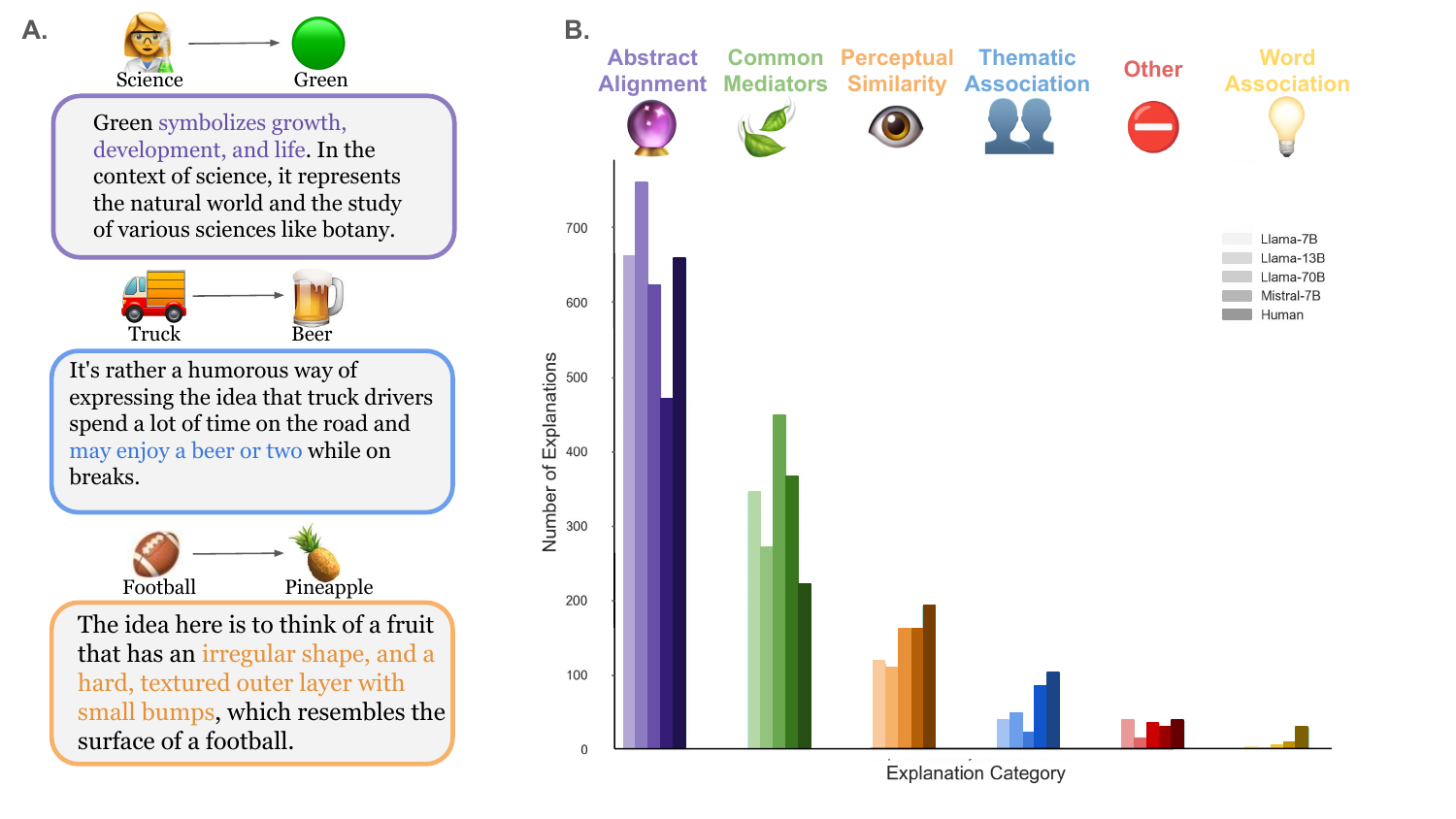}
\caption{\textbf{A}. A few examples of models' explanations from different categories \textbf{B}. Basis of cross-domain alignment, according to the model's explanations.}
\label{fig:exp_bar_example}
\end{figure*}\vspace{-1.0em}

\paragraph{Results.} Figure \ref{fig:method}B presents the M@1 results compared to the individual-level norm we define. The individual-level norm is about $17.4\%$. We can see that Flan-ul2 and Llama-7B outperform this baseline, and Llama-70B slightly underperforms it. These results indicate that some LLMs may align more closely to the average human behavior, than individual humans on this task. 

Furthermore, we investigate whether models exhibit a higher level of agreement with the most popular human answer in cases where humans also demonstrate a higher consensus on the most popular answer. To account for this, we propose a new score similar to M@1. We re-weight each statement's score by multiplying it by the number of annotators that gave that response. Then we normalize it to enable a meaningful comparison with the M@1 results. We find that the new score is higher in cases with higher human consensus, and lower in cases with lower consensus. According to this adjusted scoring method, we observe a $30\%$ improvement over the M@1 results, suggesting that LLMs, tend to agree with humans more in cases with higher consensus. This is another point of similarity between LLMs and humans.

We turn to analyze the cases where models diverge in their agreement with humans. For that, we offer a categorization of their disagreement patterns, based on manual inspection of the data: (1) Similar meaning, (2) Acceptable answer, (3) Wrong answer. To illustrate, while most participants mapped \textit{drum} to \textit{thunder}, the model mapped it to \textit{stormy}. In essence, their mappings are semantically similar, but since the score is sensitive to the form of the responses, such cases do not count as an agreement. An example of the second category is cases in which the model gives an acceptable answer that differs from the human answer. For example, the model mapped a \textit{football} to an \textit{orange}, while the typical human answer is a \textit{pineapple}. This suggests that in such cases models rely on different features than humans. Lastly, the model might generate less common or hypothetical answers that can arise from the nonsensical nature of questions in this task.

\section{Explanations}

In this section, we focus on explanations of cross-domain mappings.
We ask what type of similarity LLMs rely on when performing cross-domain mappings, and whether LLMs can reason about their responses.
For humans, the cross-domain alignment process relies on various factors, including perceptual, associative, and phonological similarity. This elicitation has shown to be valuable, as it sheds light on the underlying representations of basic concepts (LL23). Motivated by this finding we derive explanations for the LLMs mappings. We perform a qualitative and quantitative analysis of the explanations they generate for their mappings.

\paragraph{Dataset.}
For each cross-domain mapping, LL23 collected 20 distinct responses, each accompanied by a participant-provided explanation outlining the reasoning behind their choice. Subsequently, two independent raters performed annotation by classifying each response into one or more of seven predetermined categories: (1) Phonological association, (2) Perceptual similarity, (3) Word associations, (4) Common mediators, (5) Abstract alignment on certain dimensions, (6) Thematic association, and (7) Guessing (or Other; full details regarding these categories are in the Appendix). As explanations tend to have few plausible categories, the inter-rater agreement was substantial but not high, with Cohen’s kappa of $0.62$ and agreement of $61.5\%$. Most explanations, $93.5\%$, were classified into one category, and we choose to focus on this case.
The annotated explanations indicate that human explanations aren't distributed uniformly across these categories, with categories 5, 4, and 2 accounting for most explanations, with $46.1\%$ of the explanations relying on (5) Abstract alignment, alone.

\subsection{Model Explanations}\label{subsec:model_exp}

For model explanations, we use the Llama models (described in the Appendix) with the addition of Mistral-7B, a strong LLM that is specifically geared towards explanation and reasoning tasks \cite{jiang2023mistral}. We prompt the model with the question template, the human response, and a prefix for the explanation. We also add a meta-instruction that asks the model to provide a clear explanation for his answer. For this part, we use a temperature of $0.7$ to elicit a few different reasoning paths. For each statement, we have 4 templates, and for each statement we sample 4 explanations, leading to 16 explanations per statement and 1200 per model, in total.

\paragraph{Explanation Quality}

To ensure the validity of the explanations, we conduct a small-scale qualitative analysis of the model's explanations. For this analysis, we randomly sample one explanation for each of the 75 statements, focusing on Mistral-7B and Llama-13B. We manually classify\footnote{One of the authors manually annotated the data.} each explanation into (i) sensible or (ii) not sensible explanations. Overall, Mistral-7B produces more sensible explanations with $87\%$ of explanations classified as sensible, compared to Llama-13B, which scored $69\%$. Consequently, we conclude that both models exhibit a decent to high percentage of sensible explanations.

\subsection{Explanation Categories}\label{subsec:exp_cat}

We analyze model explanations by classifying them into seven predefined categories (following the classification of LL23). Initially, we use LLMs with a manually constructed prompt with one example per category and test the performance on the human-annotated explanations data. Mistral-7B performs the best at $40.8\%$, considerably below the inner-annotator agreement of $61.5\%$. To improve the results, we fine-tune a RoBERTa-large SetFit classifier with 32 examples per category, excluding 224 examples for training, which are about $16\%$ of the data \cite{tunstall2022efficient, liu2019roberta}. We deliberately choose the same number of examples per category to minimize potential classification bias. The fine-tuned classifier achieves a $60.1\%$ accuracy, comparable to the inner-annotator agreement, so we use the resultant classifier to annotate our model explanations.

In Figure \ref{fig:exp_bar_example}, we present a comparison between the distribution of explanation categories generated by LLMs and those produced by humans. To make it clearer, categories not predicted by the models are excluded from the analysis. The analysis reveals consistent trends in the dominant explanation categories employed by both humans and LLMs. Specifically, the primary categories for both groups include (5) Abstract alignment, (4) Common mediators, (2) Perceptual similarity, and (6) Thematic association, suggesting LLMs rely on similar types of similarity as humans to perform cross-domain mappings. However, this hypothesis requires further evaluation, as the link between how the LLMs explain a given mapping and the factors they rely on during the mapping process is not clear. 

Regarding categories Other, and (7) Guessing, we note that humans often use ``Guessing", whereas LLMs consistently try to provide an explanation and abstain from guessing. This difference can lead to a larger percentage of explanations assigned to the "Other" category for LLMs. Additionally, LLMs produce only a few (3) Word Association explanations and don't produce (1) Phonological Association explanations. This may suggest that LLMs are less equipped to model less frequent, end-of-distribution behaviors. Furthermore, the lower prevalence of category (3) implies that the associative strength between the source word and target domain words tends to be weaker overall. Consequently, similarly to humans, LLMs are less likely to use this type of similarity in their explanations.

These results show a strong similarity between the distribution of similarity types used by humans and LLMs as the basis of their alignment. This may suggest that they both rely on similar processes or factors while performing the mappings.

\subsection{Explanations Similarity} 

We showed that LLMs and human explanations distribute similarly across the different categories. In this section we turn to assessing whether the explanations that are produced by LLMs are similar to human explanations, rather than just falling into the same category.
To that end, we compare the pairwise similarity of LLM-human explanations with the similarity of LLM-LMM explanations. We sample one explanation per statement from the pair of LLMs or LLM and human, and for each explanation pair $(e_i,e_j)$ we compute the commonly used BERTScore metric \cite{zhang2019bertscore} across all 75 explanation pairs. We then average these scores and define the result as the similarity between the two models.\footnote{For technical reasons the value itself doesn't reflect the similarity, but can facilitate a comparison between different similarity scores.} 
We average across all similarity scores of LLM pairs and LLM-human pairs. We find that the similarity score between LLMs is $85.7\%$, while the similarity score between LLM and humans is $83.6\%$ (found significant by t-test ($p < 0.01$). This indicates that the model's explanations tend to be more similar to each other than to those of humans. A possible reason for this could be the tendency of models to provide a full and lengthy explanation, unlike most annotators' explanations, which are concise.

Moreover, we use the same setup to compare explanations across different categories. Interestingly, we see that explanations from the Perceptual Similarity category, are much more similar to each other ($87.7\%$) than other categories of explanations (4 with $84.7\%$, and 5 with $85.0\%$), suggesting that both LLMs and humans tend to rely on the same perceptual attributes in their explanations, and possibly, while performing cross-domain mappings.

\section{Discussion}
 
In this paper, we draw inspiration from the recently defined psychological task, cross-domain alignment (LL23). This task was designed to reveal how humans represent concrete and abstract concepts, by drawing insights regarding the types of similarity humans tend to rely on while performing these mappings. 
We use this task to perform a behavioral analysis of LLMs, prompting them with cross-domain mappings. 

Our results reveal that, at the population level, there is a substantial agreement between LLMs' and humans' mappings. The top-performing model predicts the most popular human answer more than $25\%$ of the times, much higher than a random baseline, and most models agree with at least one participant more than half of the time. However, as our metrics $M@K$ rely solely on word-matching and do not account for semantic equivalence, this approach might yield only a lower bound. Consequently, we defer this to future work.

We find that LLMs produce sensible explanations for the mappings. Their explanation categories distribute similarly to those of humans, which might imply that they rely on similar dimensions of similarity in their mappings. The connection between the explanations provided for the mappings and the actual factors LLMs rely on during their performance is highly non-trivial, and requires further verification.

An intriguing finding is that, similarly to humans,
LLMs tend to generate explanations that are based on perceptual similarity, although they were trained only on text. For example, mapping \textit{pineapple} to \textit{football} due to their shared texture and shape. 
This relates to the body of work that tries to assess the ability of LLMs to learn perceptual knowledge, such as visual, sensory, and phonological knowledge from text alone \cite{van2021blind,winter2023iconicity,liu2023cross,alper2023bert,marjieh2023large}. 

To conclude, we show that the seemingly nonsensical test of cross-domain mappings reveals similar patterns of behavior as presented by humans. We also use this task to assess LLM's ability to reason about these mappings, similar to how humans can ``reverse-engineer'' the reasons behind such alignments. This motivates a future study on the implications of these findings, and an examination of whether this alignment between LLMs and humans runs even deeper, namely, whether the behavioral correlates found between these tests and personality and cognitive patterns in humans can also be observed in LLMs. Given the impressive abilities presented by LLMs, several lines of work construe them as cognitive \cite{binz2023turning,kosinski2023theory,sap2022neural} and even neural \cite{huth2016natural,pereira2018toward,goldstein2022shared} models. We view this work as providing empirical underpinnings that can help map the strengths and weaknesses of this construal.










\bibliographystyle{apacite}

\setlength{\bibleftmargin}{.125in}
\setlength{\bibindent}{-\bibleftmargin}

\bibliography{CogSci}

\appendix

\section{Experimental Details}\label{app:exp_det}

\subsection{Data}

\paragraph{Domains.} For the domain-pair mappings we use the dataset created by LL23 as we describe in the paper. The domains are: 
cities, animals, instruments, beverages, colors, subjects, jobs, fruits, vehicles, sports, supernaturals, and weather.

\begin{table*}[t]
    \centering
    \resizebox{0.85\textwidth}{!}{%
        \begin{tabular}{@{}l l@{}}
            \toprule
            Category & Concepts \\
            \midrule
            Subject & History, Philosophy, Biology, Mathematics, Psychology, Science \\
            Colors & White, Red, Blue, Brown \\
            Animals & Bird, Cat, Horse, Dog, Cow, Bear, Elephant \\
            Sports & Baseball, Football, Golf, Bike, Volleyball \\
            Weather & Cloudy day, Snow, Rain, Sunny day \\
            Supernaturals & Zombie, Fairy, Wizard, Spell, Witch, Demon \\
            Jobs & Cashier, Carpenter, Doctor, Plumber \\
            Fruits & Banana, Pear, Strawberry, Apple, Grape \\
            Cities & Boston, New York, London, Beijing, Paris \\
            Instruments & Piano, Drum, Flute, Guitar, Harp \\
            Beverages & Water, Wine, Whisky, Milk, Tea, Juice \\
            Vehicles & Scooter, Bike, Truck, Motorcycle \\
            \bottomrule
        \end{tabular}%
    }
    \caption{Categories and their respective concepts.}
    \label{app:tab:categories_concepts}
\end{table*}

\subsection{Dataset.} For our experiments, we use the cross-domain mapping data collected from humans by LL23 in their experiments. Here, we describe the main steps for its construction that will be relevant for us\footnote{Full details for the data construction are also found in LL23's supplementary materials {\url{https://osf.io/tkc84/}.}}. The data was collected through Mturk from $80$ participants (40 females, 40 males, mean age = 40). After establishing 12 general domains, 32 domain pairs were selected, such that each domain is matched with two or three other domains (e.g., animal $\mapsto$ job/sports/beverage). For each domain pair, 2-3 random statements of the form:``If a(n) $x$ (source item) were a(n) $y$ (target domain), what $y$ would it be?''(e.g. ``If a \textit{doctor} (source item) were a \textit{color} (target domain), what color would it be?'')are constructed. Resulting in 75 statements, each answered by 20 participants.

\subsection{Models.} For the cross-domain mapping experiments, we select a few robust LLMs to evaluate. Flan language models \cite{wei2022finetuned} are fine-tuned on a diverse range of natural language processing (NLP) tasks and datasets, making them adaptable for various tasks through prompting. In our experiments, we use three variants: Flan-xl (3B), Flan-xxl (11B), and Flan-ul2 (20B) \cite{tay2023ul2}. Llama-2-chat \cite{touvron2023llama} is a set of large language models developed for conversational applications that undergo alignment with human preferences. We use llama-2-chat in three different sizes, referred to as Llama-7B, Llama-13B, and Llama-70B.
\subsection{Prompt Templates}
In our mapping experiment, we used the next 4 different templates. One presents the task as a question:

\begin{verbatim}
'If a $x$ (source item) were a $y$
(target domain), what would it be?
\end{verbatim}

and three as sentence compilation:
\begin{verbatim}
'If a $x$ (source item) were a $y$
(target domain), it would be a'

'If a $x$ (source item) were a $y$
(target domain), it would be an'

'If a $x$ (source item) were a $y$
(target domain), it would be'
\end{verbatim}

 We choose the indefinite article preceding $x$ and $y$ to construct a grammatical sentence. We also change the indefinite article preceding the end of the prompt to avoid directing the model toward one type of answer more than another.

For the explanations experiment, we modify the meta instruction and add an explanation prefix.

\begin{verbatim}
Answer in one word, and one word only.
Then provide a clear explanation for your answer
Question: {instruction}
Answer: {answer}
Explanation:
\end{verbatim}


\section{Re-weighted M@1 Score}
Table \ref{tbl:new_mat_scores} presents the full results of the re-weighted M@1 score, as defined in the paper. From the table, we also see that Flan-ul2 surpasses Llama-7B, and achieves $61.9\%$.

\begin{table}[t]
    \centering
    \resizebox{0.85\columnwidth}{!}{%
        \begin{tabular}{@{}lccc@{}}
            \toprule
            Model Name & RW-M@1 & @1 & Diff \\ 
            \midrule
            Flan-xl                        & 42.3 & 12.2 & 30.1   \\
            Flan-xxl                       & 38.6 & 8.1  & 30.5   \\
            Flan-ul2                       & \textbf{61.9} & 18.9 & \textbf{43.0}   \\
            Llama-7B                       &  55.4 & \textbf{24.3} & 31.1   \\
            Llama-13B                      & 55.1 & 21.6 & 29.5   \\
            \bottomrule
        \end{tabular}%
    }
    \caption{Cross-Domain Alignment scores for several models. Here, we present RW-M@1, M@1 and their differences. RW-M@1 takes into account the human consensus about the most popular answer. The higher scores of RW-M@1 indicate that the models tend to produce the most popular answer more when it is in higher consensus. 
    }
    \label{tbl:new_mat_scores}
\end{table}

\section{Explanations Categories} \label{app:exp_categories}

The alignment explanations are categorized into seven predefined categories, to reveal their underlying logic, 
following the classification established by LL23. This classification is derived from the manual examination of human annotators' rationale provided for their explanations.
In cases where an explanation fits into multiple categories, such as those observed in both our experiments and the human experiments conducted by LL23, the most suitable category is selected.

\subsection{Elaborating on Predefined Explanation Categories} 
We hereby elaborate on each of the categories\footnote{We use the examples from LL23}: \\
\textbf{Abstract Alignment.} The concepts require active abstraction to get the similarity. For example, ``If a \textit{cloudy day} were a \textit{fruit} it would be a banana'' (explanation: ``cloudy days are sad and mushy like a rotten banana.''). \\
\textbf{Perceptual Similarity.} The similarity is based on perceptual features. For example, ``If a \textit{thunder} were an \textit{instrument} it would be a drum''. \\
\textbf{Common Mediators}. Both concepts are associated with a third concept. For example, ``if a \textit{sunny day} were a \textit{fruit}, it would be an apple.'' (explanation: ``Both apple and sunny days are associated with summer.''). \\
\textbf{Word Associations.} The alignment is based on an associative relation\footnote{As this may hold true for many of the predefined categories, in LL23, the human annotators were directed to choose this category when the associations were likely formed by frequent word co-occurrence.} For example, `` if Beijing were a color, it would be red.'' (explanation: ``because it reminds me of ``red China''.'').\\
\textbf{Phonological Similarity.} The similarity is based on phonological similarity (e.g, \textit{bear} and \textit{beer}).\\
\textbf{Thematic Association.} The similarity is based on thematic (non-linguistic) association\footnote{The human annotators in LL23 were given the guideline ``If two things appear in the same context that does not necessarily have anything to do with language, it should go into this category. If the association is based on linguistic context it should go into the Word Association category.''}. For example, ``If a dog were a sport, it would be frisbee'' (explanation: ``they love games of fetch and frisbee'').\\
\textbf{Guessing.} In cases where the explanation relies on intuition or speculation. For example, `` If mathematics were a color, it would be black. '' (explanation: "I was guessing.").

\subsection{Distribution Of Explanations Categories}

In Table \ref{tab:exp_dis} we present the number of explanations for LLMs and humans and their distribution across categories. These values are the basis for Fig. \ref{fig:exp_bar_example}B. In this Table, we use the abbreviation Perceptual Sim. for Perceptual Similarity, and T-Association and W-Associations for and Thematic Association and Word Associations respectively.

\begin{table*}[t]
    \centering
    \resizebox{0.95\textwidth}{!}{%
        \begin{tabular}{@{}lcccccc@{}}
            \toprule
            Model Metric & Abstract Alignment & Common Mediators & Perceptual Sim. & T-Association & Other & W-Associations \\
            \midrule
            Llama-7B & 662 & 345 & 115 & 38 & 38 & 2 \\
            Llama-13B & 760 & 269 & 108 & 48 & 14 & 1 \\
            Llama-70B & 624 & 448 & 158 & 21 & 34 & 4 \\
            Mistral-7B & 471 & 367 & 150 & 85 & 29 & 9 \\
            Human & 659 & 221 & 192 & 102 & 38 & 29 \\
            \bottomrule
        \end{tabular}%
    }
    \caption{Distribution of explanations categories for different models.}
    \label{tab:exp_dis}
\end{table*}

\section{Explanations Similarity}
In Table \ref{tab:exp_sim}, we present the similarity between explanations of different LLMs, and human pairs, as calculated by the average BERT-Score.
Above the diagonal, we present the average similarity, and below the diagonal, we present the standard deviation. We see that LLMs tend to be more similar to one another, than to humans. This does not necessarily reveal dissimilarity between human and LLM explanations in general as the annotations explanations tend to also differentiate in style; human explanations are shorter and are not always structured as a full sentence. For instance, to explain mapping \textit{Boston} to \textit{red}, one annotator replied: ''red sox''.

\begin{table*}
    \centering
    \begin{tabular}{lccccc}
        \toprule
        & Llama-7b & Llama-13b & Llama-70b & Mistral-7b & Human \\
        \midrule
        Llama-7b & 100.0 & 87.3 & 88.6 & 87.4 & 83.6 \\
        Llama-13b & 10.4 & 100.0 & 87.7 & 86.0 & 84.1 \\
        Llama-70b & 2.2 & 1.0 & 100.0 & 87.2 & 83.7 \\
        Mistral-7b & 2.1 & 1.0 & 1.9 & 100.0 & 83.1 \\
        Human & 1.3 & 1.4 & 1 & 1.6 & 100.0 \\
        \bottomrule
    \end{tabular}
    \caption{Similarity and standard deviation between explanations of different LLMs and human pairs.}
    \label{tab:exp_sim}
\end{table*}

\section{Limitations}
Despite the insights gained from our experimentation, several limitations should be acknowledged. First, we work with strong LLMs that are not SOTA. We chose this setup as SOTA LLMs are proprietary which prevents us from knowing the process of prompting the models. Instead, we prefer to utilize the best open-source models that allow us more control of our experiment.

Second, Our score M@k is sensitive to the surface form of the responses, as we discuss in our analysis. The model might perform a semantically equivalent mapping to that of humans, but if the form of the response is different, it will not be taken into account. This concern can be mitigated by incorporating other tools that can detect semantic equivalency. For example, using word-embedding-based metrics. This is something we intend to do in future work. 

Furthermore, the human experiment data we use is of small scale (1500 data points) raising a few concerns. One concern involves the robustness of the data. Although this concern holds, the clear patterns from our experiments across different models can ease it somewhat. Additionally, the lack of social demographic details about our participants makes our results susceptible to societal biases and might not accurately reflect the phenomenon in other real-world distributions.  To address this, future work can validate our results over larger and more diverse annotator cohorts.

\end{document}